\documentclass[10pt,twocolumn,letterpaper]{article}

\usepackage[pagenumbers]{cvpr} 

\definecolor{cvprblue}{rgb}{0.21,0.49,0.74}
\usepackage[pagebackref,breaklinks,colorlinks,allcolors=cvprblue]{hyperref}

\title{Automated Interpretable 2D Video Extraction from 3D Echocardiography}

\author{
Milos Vukadinovic$^{1,2,3}$ \quad
Hirotaka Ieki$^{2,4}$ \quad
Yuki Sahashi$^{2,3}$ \quad
David Ouyang$^{2,3} \thanks{Equal contribution.}$  \quad
Bryan He$^{2,3,4}\footnotemark[1]$ \\
$^{1}$University of California, Los Angeles \quad
$^{2}$Kaiser Permanente Division of Research \\
$^{3}$Cedars-Sinai Medical Center \quad
$^{4}$Stanford University \\
{\tt milosvuk@ucla.edu, bryanhe@cs.stanford.edu}
}

\begin{document}
\maketitle
\begin{abstract}
Although the heart has complex three-dimensional (3D) anatomy, conventional medical imaging with cardiac ultrasound relies on a series of 2D videos showing individual cardiac structures. 3D echocardiography is a developing modality that now offers adequate image quality for clinical use, with potential to streamline acquisition and improve assessment of off-axis features. 
We propose an automated method to select standard 2D views from 3D cardiac ultrasound volumes, allowing physicians to interpret the data in their usual format while benefiting from the speed and usability of 3D scanning. 
Applying a deep learning view classifier and downstream heuristics based on anatomical landmarks together with heuristics provided by cardiologists, we reconstruct standard echocardiography views. This approach was validated by three cardiologists in blinded evaluation (96\% accuracy in 1,600 videos from 2 hospitals). The downstream 2D videos were also validated in their ability to detect cardiac abnormalities using AI echocardiography models (EchoPrime and PanEcho) as well as ability to generate clinical-grade measurements of cardiac anatomy (EchoNet-Measurement). We demonstrated that the extracted 2D videos preserve spatial calibration and diagnostic features, allowing clinicians to obtain accurate real-world interpretations from 3D volumes. We release the code and a dataset of 29 3D echocardiography videos \href{https://github.com/echonet/3d-echo}{https://github.com/echonet/3d-echo}.
\end{abstract}    
\section{Introduction}
\label{sec:intro}

The heart is a complex parallel pump system of two independent circulations providing deoxygenated blood to the lungs and oxygenated blood to the body. Complex anatomic structure, particularly of the right ventricle, requires holistic evaluation from multiple views.  Echocardiography, or cardiac ultrasound, is the most widespread cardiac imaging modality because it provides high-temporal resolution, no-radiation, and portable assessment of cardiac form and function \cite{mitchell_guidelines_2019}.

Early echocardiography technology was adapted from sonar and industrial flaw-detection devices, and was able to capture only 1D images (M-mode), recording the motion of cardiac structures along a single ultrasound line over time \cite{edler_history_2004}.
Rapid progress in ultrasound transducer technology has enabled high-resolution 2D imaging, which has become the standard of care in clinical echocardiography \cite{fraser_concise_2022}. Today, a typical full transthoracic echocardiography study requires a sonographer to move the probe across the patient's chest to acquire 50-100 2D videos that focus on different heart chambers and valves from different angles, a process that can take up to an hour \cite{picard_american_2011}.

Three‑dimensional echocardiography became available in the early 2000s with further improvements in ultrasound transducer technology and image processing \cite{feigenbaum_evolution_1996}. 3D volumetric datasets better capture the heart’s complex anatomy, and carry the potential to improve the speed of imaging studies, as the entire heart can be acquired as a 3D volume from a single apical chest position \cite{wu2017three}. However, unlike the transition from 1D to 2D, the shift from 2D to 3D has been more challenging, and widespread clinical adoption of 3D echocardiography has remained limited. A major barrier is the laborious reconstruction of 3D volumes into standard, clinically recognizable 2D videos for cardiologist interpretation. Currently, 2D planes are selected manually \cite{henry_three-dimensional_2022}, which is time-consuming, prone to error, and requires additional clinician training. Addressing this bottleneck would significantly increase the clinical value of 3D echocardiography, and could promote its broader adoption.

\begin{figure*}
    \centering
    \includegraphics[width=1.0\linewidth]{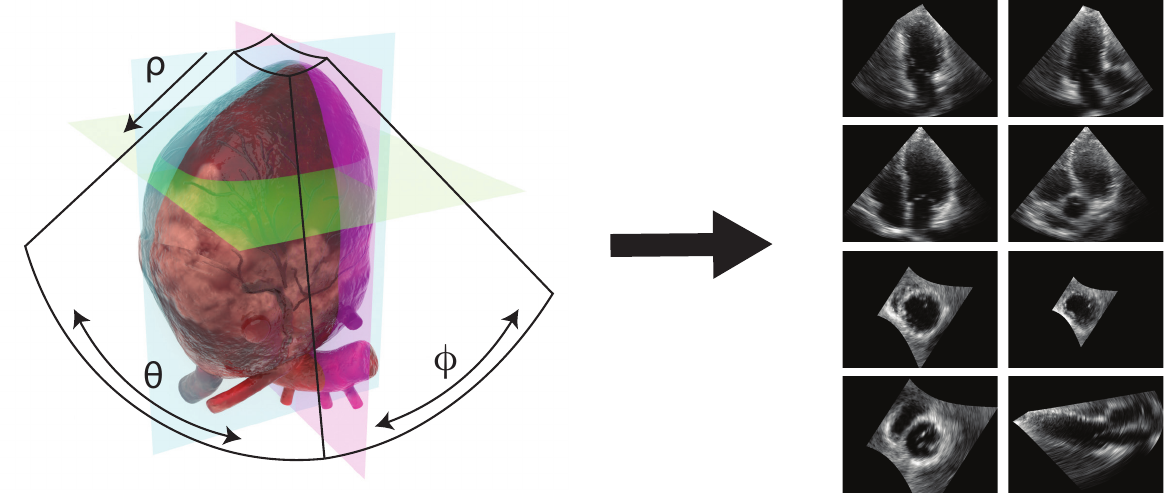}
    \caption{\textbf{Task Overview: Decomposing 3D cardiac ultrasound volumes into standard 2D images}. Left: 3D scan is acquired as a spherical pyramid. Right: Eight standard views can be extracted from the 3D scan.}
    \label{fig:fig1}
\end{figure*}

In this paper, we present a method for automated interpretable 2D view extraction from 3D echocardiography scans with comparable diagnostic quality to conventionally acquired imaging (\cref{fig:fig1}). We describe an end-to-end approach of decoding 3D volume data, a package for slicing a 3D volume and rendering 2D images along slicing planes, and an automated method using deep learning for selecting 2D planes corresponding to standard echocardiographic views. The pipeline demonstrates high-quality 2D videos upon manual inspection by cardiologists, accurate measurement of anatomic structures, and strong performance in disease detection using previously established AI models.

\section{Related Work}
\label{sec:related}

\subsection{AI in Echocardiography}
Artificial Intelligence (AI) has been used to automate various parts of echocardiography workflow \cite{myhre_artificial_2025}. Previous efforts address challenges such as probe guidance \cite{Jia_Cardiac_MICCAI2024, narang_utility_2021, Yue_2025_CVPR}, image analysis \cite{ouyang_video-based_2020, sahashi_artificial_2025}, disease prediction \cite{vrudhula_high-throughput_2024, duffy_high-throughput_2022} and structure segmentation \cite{Yue_2025_CVPR,Deng_2024_CVPR}. Several efforts have been made to build foundation models for echocardiography \cite{jiao_usfm_2024,christensen_visionlanguage_2024,amadou_echoapex_2024}, such as EchoPrime \cite{vukadinovic_echoprime_2024}, and multi-task systems such as PanEcho \cite{holste_complete_2025}. This great progress of AI in echocardiography is fueled by large 2D echocardiography datasets including publicly available datasets such as EchoNet-Dynamic \cite{ouyang_video-based_2020}, CAMUS \cite{camus}, and MIMIC-IV-ECHO \cite{johnson_mimic-iv_nodate}.

On the other hand, there are fewer applications of AI in 3D echocardiography because of 3D video dataset scarcity.  To date, no publicly available 3D video echocardiography datasets exist. The most notable effort is MITEA\cite{zhao_mitea_2023}, which provides 3D echocardiographic volumes as static images captured at systole and diastole. Nonetheless, a few research groups have assembled internal 3D video echocardiography datasets and applied machine learning to them.
Duffy et al. \cite{duffy_leveraging_2023} utilized 3D volumes to generate 2D data and test the robustness of AI model's LVEF estimation.  Pasdeloup et al. also use 3D volumes to generate 2D data but but for the purposes of training a probe guidance model
\cite{pasdeloup_real-time_2023}.
Shen et al. \cite{shen_cardiacfield_2025} developed implicit neural representation network to reconstruct a 3D cardiac volume from sequential multi-view 2DE images. 

\subsection{Slice Selection from Medical Imaging Volumes}
Human eyes cannot directly perceive 3D volumetric data, which is why selecting optimal 2D slices for visualization from a 3D volume is a well-known problem in computer vision  \cite{511,7410471,9423392,pmlr-v157-xiangwen21a}. In practice, slice selection is the most widely used in medical imaging to visualize internal structures. In CT and MRI workflows, radiologists acquire volumetric scans and then use specialized software to manually select planes of interest for viewing. However, the process of manually selecting slices is laborious, and there have been efforts to automate the slice selection process. For example, Alansary et al. \cite{frangi_automatic_2018} proposed using reinforcement learning agents to mimic experienced operators for slice selection from brain and cardiac MRI. Blansit et al. \cite{blansit_deep_2019} and Wei et al. \cite{wei_automatic_2024} developed MRI slice navigation approaches based on landmarks derived by deep learning. Similar concepts have been applied to abdominal ultrasound, where automatic view selection is typically approached in two ways: (i) reinforcement learning agents \cite{9251555}, and (ii) landmark detection models to navigate to correct views \cite{ryou_automated_2016, lorenz_automated_2018}.
 
The attempts for automated extraction of standard views from cardiac ultrasound, echocardiography, are very limited. Early work primarily focused on apical views using traditional machine learning methods. In 2014, Chykeyuk et al. \cite{chykeyuk_class-specific_2014} applied a random forest regressor to volume's voxels, using a voting scheme to predict parameters of apical planes. In the same year, Domingos et al. \cite{domingos_local_2014} combined landmark detection based on deformable models with a view classifier based on Haar features to select candidate slices and identify the best views. 

The limited research activity in this direction comes from the fact that the manual probe manipulation was both faster and yielded higher-quality images than slice selection from 3D volumes. However, advancements in transducer technology have improved the quality of 3D acquisitions. In 2022, Henry et al. \cite{henry_three-dimensional_2022} showed that 2D images reconstructed from 3D echocardiography can achieve quality comparable to conventional 2D echocardiography. This finding, coupled with the rise of AI models for echocardiography, makes automated standard view selection from 3D cardiac ultrasound increasingly feasible.

\section{Methods}
\label{sec:methods}
\subsection{Decoding 3D Echocardiograms}
A 3D echocardiography video consists of data sampled over a spherical pyramid-shaped volume (\cref{fig:fig1}).
\subsubsection{Coordinate System}
A 3D echo video is given as a grid of points $(\rho,\phi,\theta,t)$ in a spherical coordinate system with non-standard angle definitions. Namely, $\rho$ is the distance of the point from the origin, $\phi$ is the azimuthal angle in the x-z plane measured from the x axis, $\theta$ is an elevation angle from the x-z plane, and $t$ is the frame number. Throughout this text we continue calling it the spherical coordinate system. Formulas for conversion to cartesian coordinates are:

\begin{equation}
    \begin{aligned}
    x &= \rho \cos(\phi) \cos(\theta) \\
    y &= \rho \sin(\theta) \\
    z &= \rho \sin(\phi) \cos(\theta)
    \end{aligned}
    \label{eq:cartesian_conversion}
\end{equation}
\subsubsection{Decoding Volumes from Patient Records}
First, we obtain the shape of the 3D volume $(\rho, \phi, \theta)$ from the DICOM metadata. The actual voxel intensity values are then stored in the metadata as a raw byte stream, as illustrated in  \cref{fig:sup_fig1} (Supplementary). Then, to interpret the volume in physical space, we extract the ranges over which $\rho$, $\phi$, and $\theta$ are defined, namely:
$(\phi_{\min}, \phi_{\max})$,
$(\rho_{\min}, \rho_{\max})$,
$(\theta_{\min}, \theta_{\max})$.
\subsubsection{Point Cloud Representation}
After decoding, we have a grid of voxel intensities $V \in \mathbb{R} ^{I \times J \times K \times T}$ and we have bounds on which the volume is defined.
\begin{equation}
     B = \begin{bmatrix}
     \rho_{min} & \rho_{max} \\
     \phi_{min} & \phi_{max} \\ 
     \theta_{min} & \theta_{max} 
     \end{bmatrix}
     \label{eq:bound_matrix}
\end{equation}

To get the Cartesian coordinates of each grid entry $(i,j,k,t)$, we perform linear sampling within the given bounds for each axis.
$C \in \mathbb{R}^{I \times J \times K \times 3} $

\begin{equation}
    \begin{aligned}
    C[i,j,k,0] = \rho_{min} + i \frac{(\rho_{max}-\rho_{min})}{I-1} \\
    C[i,j,k,1] = \phi_{min} + j \frac{(\phi_{max}-\phi_{min})}{J-1} \\
    C[i,j,k,2] = \theta_{min} + k \frac{(\theta_{max}-\theta_{min})}{K-1}
    \end{aligned}
    \label{eq:coordinates_sampling}
\end{equation}

This way we achieve a point-cloud representation, because for a given index $idx \in I \times J \times K$ at time $t \in T$ we can obtain its coordinates from ${C}[idx]$ and its voxel intensity from ${V}[idx,t]$.
\subsubsection{Reproducibility and Dataset Release}
All 3D echocardiograms in our dataset were acquired with a Philips EPIQ CVx system. Because the 3D data is stored in private DICOM fields, we describe a decoding process, and publish the code on github. To support reproducibility, we also release the dataset of 29 3D echocardiography videos acquired from four consenting volunteers from our research team. We hope that these efforts will promote transparency and accelerate the development of machine learning methods for 3D echocardiography.
\begin{figure*}
    \centering
    \includegraphics[width=1.0\linewidth]{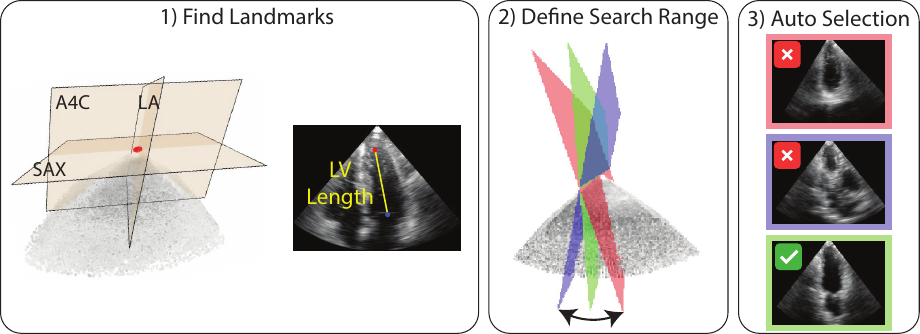}
    \caption{\textbf{Overview of the proposed view extraction method}: 1) A segmentation model localizes key landmarks (A4C, LA, SAX, and LV length). 2) A plane search range is defined using cardiologist-provided heuristics and the detected landmarks. 3) A view classifier performs the search and automatically selects the best views.}
    \label{fig:fig2}
\end{figure*}

\subsection{Obtaining a 2D slice from a volume}
\subsubsection{Parametrizing 2D slices}
To obtain a 2D slice from a 3D video, we begin by defining the plane that determines where pixel values are sampled. The same plane is applied across all time frames, making the temporal dimension independent of the slicing operation. We define three parametrizations of the plane and switch between them depending on convenience. 

Parametrization 1 (Point-Normal Form) is used to derive planes that pass through segmentation landmarks. Parametrization 2 (Angle-Distance Form) is used for plane search, because it has the  fewest degrees of freedom and corresponds to adjustments of the probe in standard 2D echocardiography. Parametrization 3 (Parametric Form) is used to sample the 2D image corresponding to a plane, as it allows straightforward generation of uniformly distributed points on the plane.
\newline
\newline
\textbf{Parametrization 1: Point-Normal Form}
\begin{equation}
    \mathcal{P}_1: P = (p_x, p_y, p_z) \in \mathbb{R}^3, \quad 
    n=(n_x, n_y, n_z) \in \mathbb{R}^3
\end{equation}
The plane is specified by a point $P$ lying on it and a normal vector $n$. A point $Q \in \mathbb{R}^3$ lies in the plane if
\begin{equation}
    n\cdot (Q-P) =0
    \label{eq:plane1}
\end{equation}
\textbf{Parametrization 2: Angle-Distance Form}
\begin{equation}
    \mathcal{P}_2: (d, \phi_n, \theta_n)
\end{equation}
where $\phi$ is the azimuthal angle of the normal vector, $\theta$ is the elevation angle, and  $d$ is the signed distance from the origin to the plane along its normal. 
To obtain angle-distance form from point-normal form we can use these equations:
\begin{equation}
    d=\frac{n p}{||n||}
\end{equation}
\begin{equation}
    \phi_n = \mathrm{atan2}(n_z,n_x)
\end{equation}
\begin{equation}
    \theta_n = \mathrm{asin}(n_y)
\end{equation}

after which the angles are converted to degrees.
\textbf{Parametrization 3: Parametric Form}
\begin{equation}
Q(s,t) = P + su + tv, \quad s,t \in \mathbb{R}
\end{equation}
where $u$ and $v$ are two orthogonal vectors lying in the plane, defined as: 
\begin{equation} u = \begin{cases} n \times [1,0,0] & \text{if } n \neq [1,0,0] \\
n \times [0,1,0], & $ \text{otherwise}$
\end{cases}
\end{equation}
$$ v = n \times  u$$

\subsubsection{Defining the Sampling Grid}
\label{subsubsec:sampling_grid}
The sampling grid is defined by the intersection of the cutting plane with the 3D volume. Using the plane’s parametric form (parametrization 3), we determine the bounded region by projecting the volume’s point cloud onto the plane’s basis vectors $u$ and $v$ and computing the minimum and maximum values of $s$ and $t$.
\begin{equation}
s_{min} = \mathrm{min}( (\hat{C} - P) u) \quad s_{max} = \mathrm{max}((\hat{C} - P) u)
\end{equation}
\begin{equation}
    t_{min} = \mathrm{min}((\hat{C} - P) v) \quad t_{max} = \mathrm{max}((\hat{C} - P) v)
\end{equation}
Here, $C$ denotes the coordinate matrix in spherical space $(\rho,\phi,\theta)$, and $\hat{C}$ denotes the corresponding Cartesian coordinates obtained by applying the spherical-to-Cartesian transformation.

We then uniformly sample values from $[s_{\min}, s_{\max}]$ and values from $[t_{\min}, t_{\max}]$ to construct the sampling grid.
\begin{equation}
    \hat{G}(i,j) = P + s_iu +t_jv
\end{equation}
Each entry $\hat{G}(i,j)$ corresponds to the pixel location $(i,j)$ in the 2D slice and stores the Cartesian coordinates $(x,y,z)$ of the corresponding 3D point in the volume from which the intensity value will be sampled. 

\subsubsection{Interpolating pixel intensities}
We have defined a sampling grid, but because we are working with a point cloud, we need to interpolate pixel intensities. We perform the interpolation in the spherical coordinate system. First, we convert a sampling matrix $\hat{G}$ into the spherical coordinate system and obtain $G$. Then we perform trilinear interpolation to sample points specified by $G$ from a 3D rectilinear grid defined by the matrix of bounds $B$ and matrix of values $V$. The output is a $(h_{\mathrm{pix}} \times w_{\mathrm{pix}})$ array representing the 2D slice image.

Finally, once a slice is rendered as an image, we must ensure the correct viewpoint and spatial resolution (centimeters per pixel). This introduces extra hyperparameters: centimeters per pixel, horizontal flip, vertical flip, and in-plane rotation. These are view-specific constants rather than learned parameters, so we defer their detailed description to the Supplementary Material and provide their values in the released code.
\begin{table*}[htbp]

    \centering
    \begin{tabular}{lccc}
    \hline
    Parameter & $\boldsymbol{d}$ & $\boldsymbol{\phi}$ & $\boldsymbol{\theta}$ \\
    \hline
    A2C & $ d_{LA} $ & $\phi_{LA}$ & ($\theta_{LA}$, $\theta_{LA}+30$)  \\
    A3C & $d_{LA} $   & $\phi_{LA}$  & ($\theta_{A2C}-60, \theta_{A2C}-15$) \\
    A4C & $d_{A4C}$ & $\phi_{A4C}$ & $\theta_{A4C}$ \\
    A5C & $d_{A4C}$ & $\phi_{A4C}$ & ($\theta_{A4C}+10, \theta_{A4C}+35$) \\
    SAX apex &  ($d_{SAX} + 0.10 \cdot \l_{LV}, d_{SAX} + 0.20\cdot\l_{LV})$ & $\phi_{SAX}$  & $\theta_{SAX}$  \\
    SAX PAP  &  ($d_{SAX} + 0.40\cdot\l_{LV}, d_{SAX} + 0.50\cdot\l_{LV})$  & $\phi_{SAX}$ & $\theta_{SAX}$   \\
    SAX MV & ($d_{SAX} + 0.75\cdot\l_{LV}, d_{SAX} + 0.80\cdot\l_{LV})$ & $\phi_{SAX}$  & $\theta_{SAX}$ \\
    PLAX & $d_{LA} $   & $\phi_{A3C}$  & $\theta_{A3C}$  \\
    \hline
    \end{tabular}
    \caption{\textbf{Plane search ranges for all standard echocardiography views.}}
    \label{tab:PlaneParams}

\end{table*}
\subsection{Finding Standard Views}
Now that we are able to cut the 3D volume across arbitrary planes and render the slices as images, we want to find what planes correspond to standard echocardiographic views. Eight standard echocardiographic views are:
Apical 2 Chamber (A2C), Apical 3 Chamber (A3C), Apical 4 Chamber (A4C), Apical 5 Chamber (A5C), Parasternal Long AXis (PLAX), parasternal Short AXis level of apex (SAX apex), parasternal Short AXis level of PAPillary muscles (SAX PAP) and parasternal Short AXis level of Mitral Valve (SAX MV). The overview of the proposed standard view extraction method is illustrated in \cref{fig:fig2}.

\subsubsection{Landmark Localization}
We start from a 3D volume represented as a point cloud and first localize anatomical landmarks. We aim to identify four landmarks:
\begin{itemize}
\item $\mathcal{P}_{A4C} : (d_{A4C},\phi_{A4C},\theta_{A4C})$ A4C plane at the apex point in angle-distance form (parametrization 2)
\item $l_{LV}$: left ventricle length
\item $\mathcal{P}_{SAX} : (d_{SAX},\phi_{SAX},\theta_{SAX})$ short axis plane at the apex point
\item $\mathcal{P}_{LA} : (d_{LA},\phi_{LA},\theta_{LA})$ long axis plane at the apex point
\end{itemize} 
Since the 3D scan is acquired from the apical position, A4C plane at the apex point is easy to specify. It corresponds to looking straight down the probe axis, with angle-distance form $(d_{A4C}, \phi_{A4C}, \theta_{A4C}) = (0,0,90)$. Next, we render the image in the A4C plane and use EchoNet-Dynamic segmentation model \cite{ouyang_video-based_2020} to find apex-to-base line to obtain $l_{LV}$. This segmentation also provides the apex coordinates $P_{apex}$ and the apex-to-base vector $v_{apex}$. With these, we define the short-axis plane $\mathcal{P}_{SAX}$ in point–normal form ($P=P_{apex}, n = v_{apex}$), and subsequently convert it to angle–distance form. Finally, $\mathcal{P}_{LA}$ can be defined in point–normal form using the apex point and a vector orthogonal to the normal vectors of $\mathcal{P}_{SAX}$ and $\mathcal{P}_{A4C}$. As before, we convert from point–normal to angle–distance form to facilitate defining search ranges with minimal degrees of freedom.

\subsubsection{Defining Search Ranges}
 In routine transthoracic echocardiography (TTE), sonographers follow general guidelines for transitioning between views. These are not strict rules, as cardiac anatomy varies between individuals, but rather suggested search ranges. Sonographers move the probe within these search ranges and pick the best view by visually confirming that the target view's characteristic anatomical structures are visible. Assuming we can localize the A4C, SAX, and LA planes as well as LV length (corresponding to our landmarks), the transition rules sonographers follow can be summarized as

\begin{itemize}

\item A2C: Rotate the transducer $0–30^\circ$ clockwise from the long-axis plane.

\item A3C: From A2C, rotate the transducer $15–60^\circ$ counterclockwise.

\item A5C: From A4C, tilt the probe anteriorly by $10–35^\circ$.

\item SAX Apex: Defined within the SAX plane at $10–20\%$ of LV length.

\item SAX Papillary: Defined within the SAX plane at $40–50\%$ of LV length.

\item SAX Mitral Valve: Defined within the SAX plane at $75–80\%$ of LV length. 

\item PLAX: Same as A3C, but with the viewpoint rotated by approximately $70^\circ$ (in-plane rotation).

\end{itemize}

Motivated by this practice, we define analogous search ranges for each view based on our landmarks, as summarized in \cref{tab:PlaneParams}.

\subsubsection{Auto Selection}
We used the echocardiography view classifier (from \cite{vukadinovic_echoprime_2024}) to identify the best candidate plane within each search range. Each plane is parameterized by $(d, \phi, \theta)$, where the corresponding search ranges are defined as intervals, e.g., $(d_{\min}, d_{\max}), (\phi_{\min}, \phi_{\max}) , (\theta_{\min}, \theta_{\max})$. Within each range, we exhaustively sampled candidate planes by incrementing each parameter in steps of 1 unit. Candidate slices were rendered as images using a frame at end-diastole. The view classifier then assigned probabilities for the target view, and the slice with the highest probability was selected as the final choice.

\section{Experiments}
\label{sec:experiments}

\subsection{Experimental Setup}
Our experiments aim to evaluate whether views automatically extracted from 3D volumes are comparable to standard echocardiographic views acquired by moving the ultrasound probe across the patient's chest, both in appearance and informational content. For all experiments we utilize 3D Echocardiography volumes from two different medical institutions. The first dataset is from Cedars-Sinai Medical Center (CSMC) containing 1606 3D volumes, and the second dataset is from Stanford Healthcare (SHC) containing 120 3D volumes, where each volume comes from a different echocardiography study. The proposed method is applied to extract $8$ standard echocardiography views from each volume.
We evaluate the quality of the extracted views using three experiments: manual assessment by cardiologists, AI-enabled disease detection and tracing of structural measurements.

\subsection{Cardiologists Assessment}
Three expert cardiologists collectively reviewed $1600$ extracted videos from $200$ 3D volumes ($8$ videos per volume, $100$ volumes from CSMC and $100$ volumes from SHC). Cardiologists first assessed the quality of each video, noting cases with unrecognizable views, significant noise, or marked artifacts. Then after the quality check, cardiologists assigned the best matching view, blinded to the labels assigned by our method.

\begin{figure}
    \centering
    \includegraphics[width=1.0\linewidth]{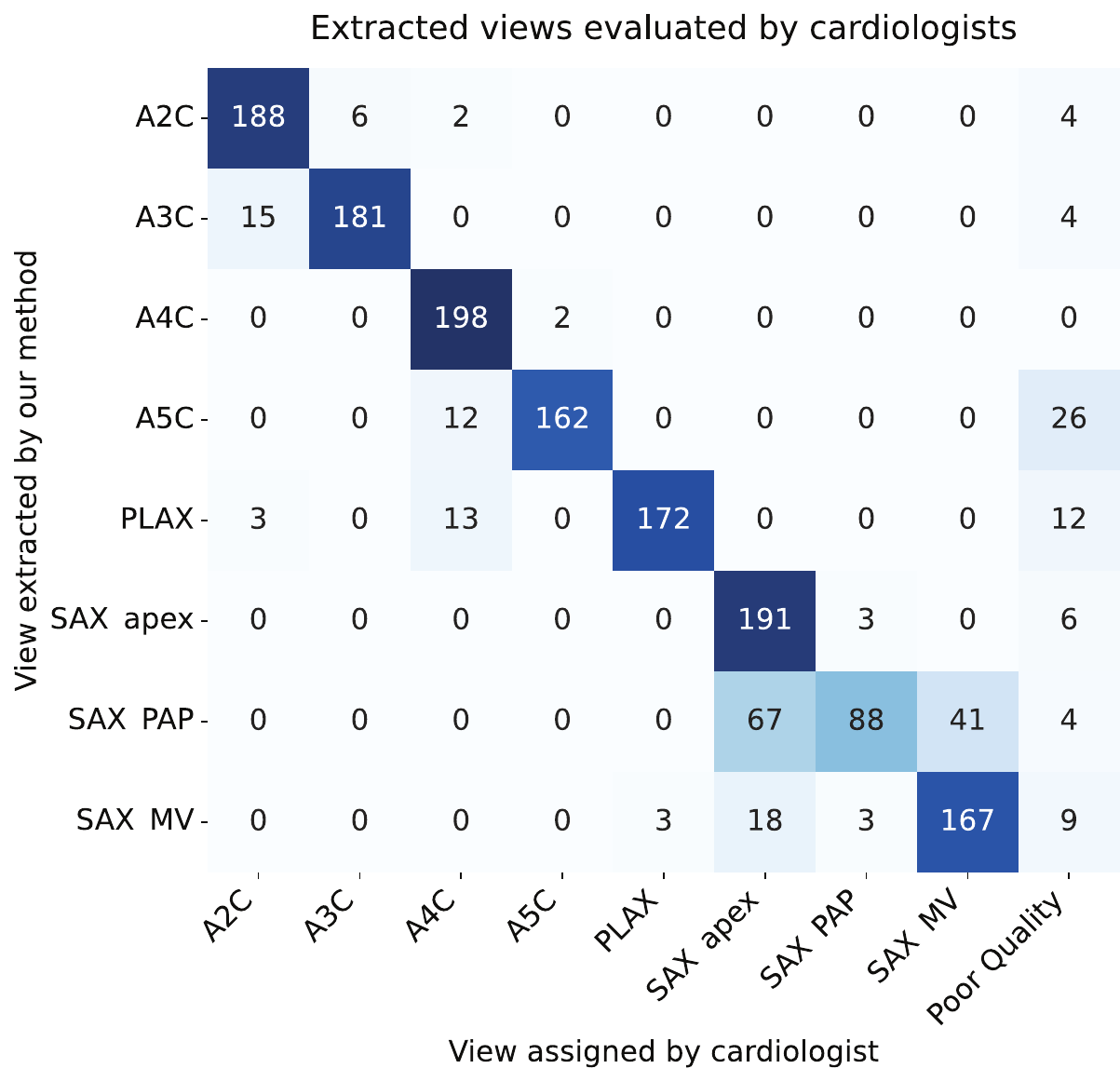}
    \caption{\textbf{Results from Cardiologists Assessment.} Cardiologists were asked to assess quality and view-correctness of $1600$ extracted videos (8 views × 100 videos per view × 2 institutions).}
    \label{fig:fig3}
\end{figure}

Agreement between cardiologist-assigned views and our method’s outputs is reported as per-view accuracies ( \cref{tab:per_view_accuracy}) and a confusion matrix (\cref{fig:fig3}). Overall, 96.5\% of the videos in CSMC and 95.5\% in SHC were labeled as good quality. On CSMC dataset cardiologist assessed 84.8\% of the views as both good quality and correct view and on SHC 83.6\%. The highest per-view accuracy was for A4C ($99.0\%$). This is expected because all volumes in our dataset are acquired from the apical position, making it easier to find A4C cutting plane.
The lowest per-view accuracy was for SAX PAP (44.0\%). However, as shown in the confusion matrix, cardiologists consistently recognized SAX PAP as a short-axis ($SAX$) view, with disagreement arising only in distinguishing the specific level (apical, papillary, or mitral valve). In particular, $98\%$ of SAX PAP views were labeled as high-quality short-axis views.

\begin{table}[h]
\centering
\caption{\textbf{Per-view accuracies}: The proportion of videos in which cardiologists confirmed the extracted view was correct.}
\begin{tabular}{lcc}
\toprule
\textbf{View} & \textbf{Accuracy (\%)} \\
\midrule
A2C        & 94.0 \\
A3C        & 90.5 \\
A4C        & 99.0 \\
A5C        & 81.0 \\
PLAX       & 86.0 \\
SAX apex  & 95.5 \\
SAX PAP   & 44.0 \\
SAX MV    & 83.5 \\
\bottomrule
\end{tabular}
\label{tab:per_view_accuracy}
\end{table}
\begin{table*}[htbp]
\centering

\begin{tabular}{lccc}
\hline
\multicolumn{4}{c}{\textbf{EchoPrime}} \\
\toprule
& \multicolumn{2}{c}{\textbf{3D}} &\textbf{2D}\\ 
\cmidrule(lr){2-3} \cmidrule(lr){4-4}
Metric & \textbf{Random Cuts} & \textbf{Our Method} & \textbf{Benchmark}  \\
\midrule
Ejection Fraction R2 & 0.42 (0.36--0.48) & \textbf{0.74 (0.71--0.77)} & 0.83 (0.81--0.85) \\
Ejection Fraction MAE & 7.56 (7.19--7.94) & \textbf{5.34 (5.10--5.57)} & 4.28 (4.09--4.47) \\
RV Systolic Dysfunction AUC & 0.81 (0.75--0.85) & \textbf{0.88 (0.84--0.92)} & 0.95 (0.93--0.97) \\
LV Dilation AUC & 0.83 (0.77--0.88) & \textbf{0.87 (0.82--0.92)} & 0.93 (0.90--0.95) \\
LA Dilation AUC & 0.76 (0.69--0.82) & \textbf{0.86 (0.80--0.91)} & 0.93 (0.89--0.96) \\
RV Dilation AUC & 0.77 (0.69--0.84) & \textbf{0.79 (0.71--0.86)} & 0.92 (0.85--0.97) \\
RA Dilation AUC & 0.83 (0.73--0.91) & \textbf{0.90 (0.83--0.96)} & 0.98 (0.97--0.99) \\
Mitral Regurgitation AUC & 0.75 (0.70--0.79) & \textbf{0.85 (0.82--0.88)} & 0.92 (0.90--0.94) \\
Mitral Stenosis AUC & 0.55 (0.49--0.68) & \textbf{0.92 (0.78--0.98)} & 0.97 (0.96--0.99) \\
Aortic Regurgitation AUC & 0.53 (0.45--0.60) & \textbf{0.77 (0.72--0.83)} & 0.93 (0.90--0.96) \\
Aortic Stenosis AUC & 0.61 (0.54--0.68) & \textbf{0.93 (0.88--0.97)} & 0.98 (0.95--0.99) \\
\bottomrule
\\
\hline
\multicolumn{4}{c}{\textbf{PanEcho}} \\
\toprule
& \multicolumn{2}{c}{\textbf{3D}} & \textbf{2D} \\
\cmidrule(lr){2-3} \cmidrule(lr){4-4}
Metric & \textbf{Random Cuts} & \textbf{Our Method} & \textbf{Benchmark} \\
\midrule
Ejection Fraction R2 & -0.13 (-0.22--0.04) & \textbf{0.63 (0.59--0.67)} & 0.56 (0.51--0.61) \\
Ejection Fraction MAE & 11.52 (11.06--11.98) & \textbf{6.18 (5.89--6.47)} & 6.62 (6.30--6.94) \\
RV Systolic Dysfunction AUC & 0.63 (0.52--0.73) & \textbf{0.85 (0.77--0.92)} & 0.90 (0.84--0.95) \\
LV Dilation AUC & 0.64 (0.58--0.70) & \textbf{0.89 (0.85--0.93)} & 0.90 (0.86--0.93) \\
LA Dilation AUC & 0.53 (0.45--0.61) & \textbf{0.80 (0.75--0.85)} & 0.82 (0.78--0.85) \\
RV Dilation AUC & 0.56 (0.46--0.66) & \textbf{0.81 (0.73--0.87)} & 0.86 (0.79--0.91) \\
RA Dilation AUC & 0.70 (0.60--0.80) & \textbf{0.92 (0.85--0.97)} & 0.94 (0.90--0.97) \\
Mitral Regurgitation AUC & 0.60 (0.55--0.65) & \textbf{0.84 (0.81--0.87)} & 0.85 (0.82--0.88) \\
Mitral Stenosis AUC & 0.76 (0.61--0.90) & \textbf{0.95 (0.92--0.99)} & 0.92 (0.84--0.99) \\
Aortic Regurgitation AUC & 0.54 (0.47--0.62) & \textbf{0.66 (0.59--0.73)} & 0.73 (0.66--0.80) \\
Aortic Stenosis AUC & 0.55 (0.47--0.63) & \textbf{0.76 (0.70--0.82)} & 0.92 (0.87--0.95) \\
\bottomrule
\end{tabular}

\caption{\textbf{Performance of AI echocardiography models on automatically extracted views.} \emph{Random Cuts} column reports metrics on randomly selected slices from $3D$ volumes without guided slice selection. \emph{Our Method} column shows metrics on videos extracted from $3D$ volumes using our approach. \emph{Benchmark} column reports metrics on original 2D videos acquired by sonographers and serves as the upper bound.  }
\label{tab:EchoPrimeMetrics}
\end{table*}

\subsection{AI-enabled echocardiography interpretation}
In the next experiment, we evaluated whether the extracted views could be used for accurate interpretation of echocardiography exams. We employed EchoPrime \cite{vukadinovic_echoprime_2024} and PanEcho \cite{holste_complete_2025},  previously developed models for various echocardiography tasks, and inferenced them on videos extracted from 3D volumes using our method. We averaged predictions across all views in the study to get study-level predictions and compared it with the ground-truth labels from clinical databases on $10$ key echocardiographic tasks.

For regression tasks we report the coefficient of determination ($R2$)  and mean absolute error (MAE), while for binary tasks we report area under the receiver operating characteristic curve (AUC). For all binary tasks, the ground truth label is positive if the severity of condition is moderate or greater. Results presented are based on $1606$ studies from CSMC. $120$ studies from SHC were not included due to insufficient ground-truth labels in the clinical databases for reliable statistical evaluation.

We compared our approach against the unguided slice selection method, Random Cuts. This baseline evaluates performance without guided slice selection by choosing eight random cutting planes per volume. For reference, we also report Benchmark metrics, which use the original 2D videos acquired directly by sonographers through probe manipulation rather than views extracted from 3D volumes.

The results are shown in \cref{tab:EchoPrimeMetrics}.
For left ventricular ejection fraction, EchoPrime achieved $5.34$ MAE on views extracted using our method versus $7.56$ MAE on views extracted using random cuts. On binary metrics, EchoPrime had an average $0.86$ AUC with our method, versus $0.72$ AUC with random cuts. 
Because EchoPrime was also trained on dataset from CSMC, raising the possibility of inflated performance if our test set overlapped with EchoPrime's train set, we additionally evaluate PanEcho, trained on a dataset from Yale New Haven Health System.
On estimating the ejection fraction task, PanEcho achieves $6.18$ MAE on views extracted using our method versus $11.52$ MAE on views extracted using random cuts. On binary metrics, PanEcho had an average $0.83$ AUC with our method, versus $0.61$ AUC with random cuts.

An important observation is that PanEcho’s performance on 3D extracted views using our method consistently falls within the confidence intervals of its performance on sonographer-acquired 2D videos (Benchmark). This finding supports our hypothesis that 2D views automatically extracted from 3D volumes can be as good as sonographer acquired 2D views for the purposes of echocardiography interpretation. In practice, this would allow for keeping the same accuracy as standard 2D workflow while benefiting the speed of 3D acquisition.

\begin{figure*}[htbp]
    \centering
    \includegraphics[width=1.0\linewidth]{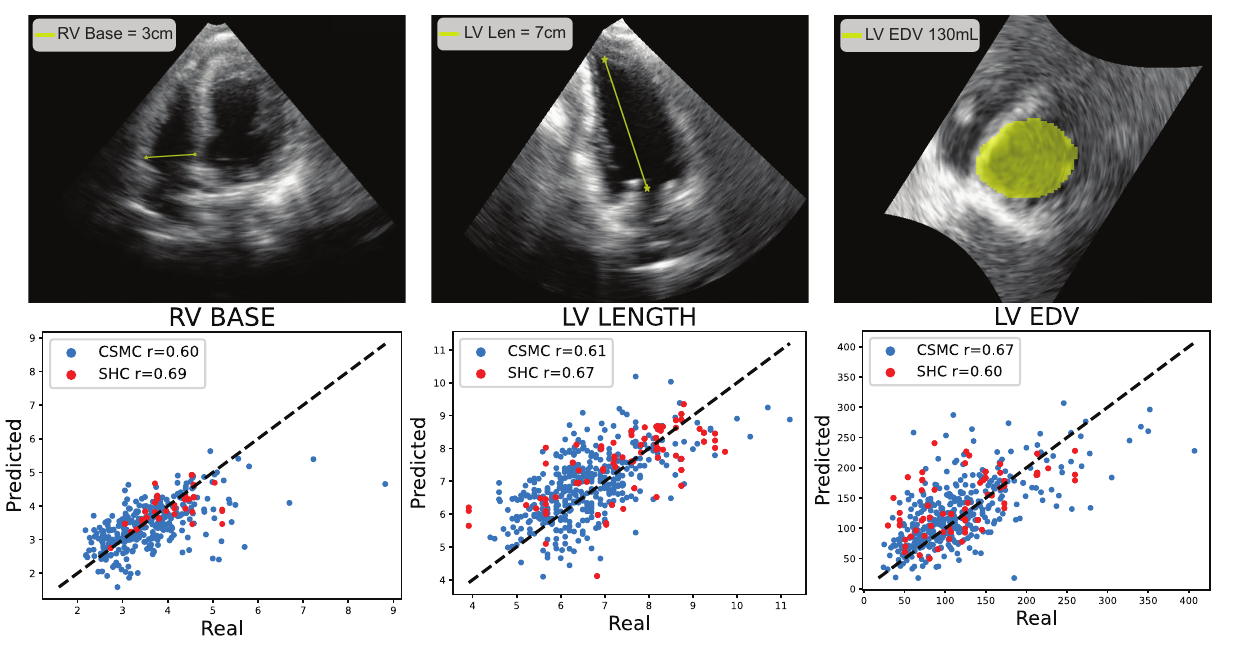}
    \caption{\textbf{Measurement Tracing on Extracted Views.} Left: Visualization of the traced measurements. Right: Scatterplots with correlation coefficients against ground truth across two datasets.}
    \label{fig:fig4}
\end{figure*}  
\subsection{Structural Measurement Tracing}
When analyzing echocardiography studies, clinicians often measure lengths and volumes directly from the videos, because obtaining these values in centimeters provides critical information for diagnosis. That is why we designed an experiments to assess how accurately can we trace measurements from the extracted videos, i.e. we test if extracted views preserve the spatial calibration.

We focused on three measurements: RV-Base (right ventricle base length), LV-Length (left ventricular apex-to-base length) and EDV (left ventricular end-diastolic volume) (\cref{fig:fig4} Left). We utilized three different previously published AI models for measurement calculation from echocardiography videos.
EchoNet-Measurements \cite{sahashi_artificial_2025} was used to calculate RV-Base and EchoNet-Dynamic \cite{ouyang_video-based_2020} was used to calculate LV-Length. EchoNet-Pediatric \cite{reddy_video-based_2023} was used to get left ventricular areas from short axis views which can then, together with left ventricle length provided by EchoNet-Dynamic, be used to estimate EDV using the formula ($\sum_{i=1}^{N} \frac{LV\-Length}{N} Area[i]$).
Measurements predicted by these models are then compared to the ground truth values from the clinical databases. Pearson correlation coefficient $r$ is reported.

Scatterplots and correlation coefficients for all three measurements accross two datasets are shown in \cref{fig:fig4}. The correlation between ground-truth measurements (real) and AI predicted measurements (predicted) was $r=0.60$ on $330$ RV Base samples from CSMC and $r=0.69$ on $67$ RV Base samples from SHC. For LV Length, the correlation was $r=0.61$ on $409$ CSMC samples and $r=0.67$ on $196$ SHC samples. For LV EDV, $r=0.67$ on $367$ CSMC samples, and for LV EDV on SHC, $r=0.60$ on $69$ samples. Altough some variance in predictions is expected due to our reliance on automated AI-based tracing rather than manual human tracing, these results are strong and demonstrate that the extracted views preserve spatial calibration.

\section{Conclusion}
\label{sec:conclusion}

In this paper, we propose an approach for automated extraction of interpretable 2D videos from 3D  echocardiography scans. We introduced a mathematical framework for slicing arbitrary planes within the volume and rendering them as images, together with an algorithm for selecting standard echocardiographic views guided by the view classifier and anatomical landmarks.
Extensive experiments confirmed the quality of the extracted views: cardiologists rated them as high-quality and clinically informative, and using AI models for echocardiography we demonstrated that they can effectively be used for cardiac disease detection. Our method aims to streamline the traditionally lengthy and skill-dependent process of maneuvering an ultrasound probe to acquire standard echocardiographic views. Instead, a single 3D scan acquired from the apical position can be computationally converted to standard 2D views. This allows physicians to interpret the data in their usual format while leveraging the speed and ease of 3D scan acquisition.

{
    \small
    \bibliographystyle{ieeenat_fullname}
    \bibliography{main}
}

\clearpage
\setcounter{page}{1}
\maketitlesupplementary

\section{View Specific Hyperparameters for correct Spatial Orientation}
After selecting the slice containing an interpretable 2D video from a 3D echocardiography volume, we still need to adjust spatial parameters so that the resulting videos resemble those typically interpreted by physicians.
\subsection{Spatial Resolution}
Given a sampling grid, as defined in \cref{subsubsec:sampling_grid} the physical dimensions of the slice (in centimeters) are given by
$$ w = || (s_{max}-s_{min}) u ||$$
$$ h = || (t_{max} - t_{min}) v ||$$
A desired spatial resolution (in cm/pixel) is then specified to determine how many pixels to sample along the width and height.
$$ w_{pix} \times h_{pix} = \frac{w}{cm/pix} \times \frac{h}{cm / pix}$$

We set the resolution parameter ($cm / pix$) to the most common value observed in the dataset for each view.
\subsection{Horizontal Flip}
A horizontal flip reverses the readout direction of the sampling grid along the vector $u$. I.e. it replaces $u$ with $-u$.
\subsection{Vertical Flip}
A vertical flip reverses the readout direction of the sampling grid along the vector $v$. I.e. it replaces $u$ with $-u$.
\subsection{Rotation Angle}
The rotation angle is applied directly to the rendered image, rotating it in the image plane by the specified angle.
\begin{figure*}
    \centering
    \includegraphics[width=1.0\linewidth]{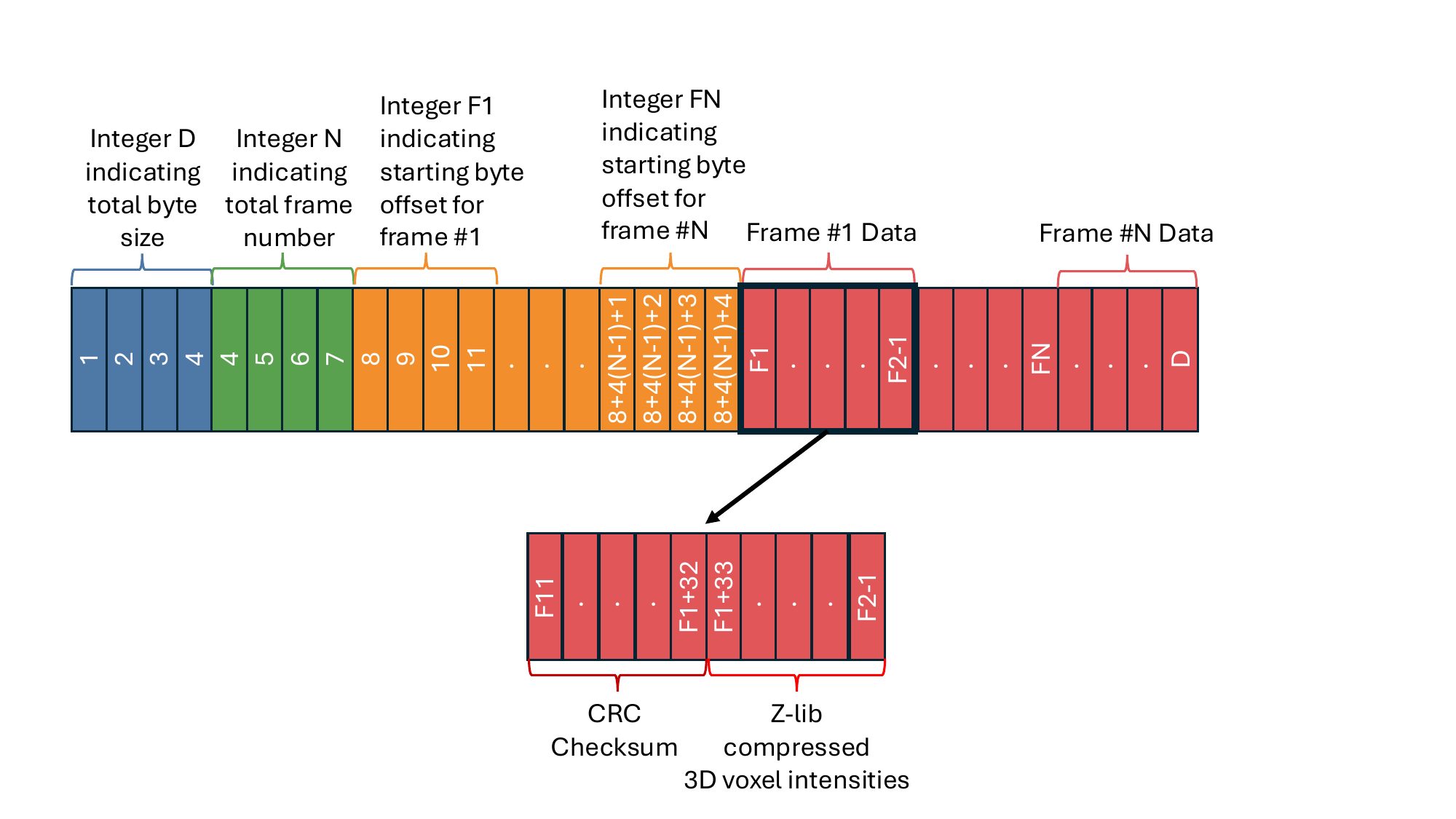}
    \caption{\textbf{Decoding Bytes to Voxel Intensities.} The first 4 bytes represent an integer indicating the total data size in bytes, followed by 4 bytes specifying the total number of frames. Next, a series of 4-byte integers (one for each frame) provide the starting byte offset of each frame within the stream. For each frame, the first 32 bytes are a CRC checksum, and the remaining bytes are a zlib-compressed 3D volume voxel intensities.}
    \label{fig:sup_fig1}
\end{figure*}

\end{document}